\documentclass{article} % For LaTeX2e
\usepackage{colm2024_conference}

\usepackage{microtype}
\usepackage{hyperref}
\usepackage{url}
\usepackage{booktabs}
\usepackage{multirow}
\usepackage{multicol}
\usepackage{makecell}
\usepackage{graphicx}
\usepackage{wrapfig}
\usepackage{amsthm}
\usepackage{xcolor,colortbl}
\usepackage{amssymb}
\usepackage{pifont}
\usepackage{fdsymbol}
\usepackage{color, colortbl}
\newtheorem{definition}{Definition}
\newcommand{\kn}{\raisebox{-0.9mm}{\includegraphics[width=4mm]{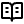}}}
\newcommand{\work}{\raisebox{-0.8mm}{\includegraphics[width=4mm]{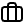}}}
\newcommand{\world}{\raisebox{-0.8mm}{\includegraphics[width=4mm]{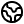}}}
\newcommand{\modal}{\raisebox{-0.8mm}{\includegraphics[width=4mm]{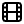}}}
\newcommand{\nn}{\raisebox{-0.8mm}{\includegraphics[width=4mm]{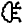}}}
\definecolor{ao}{rgb}{0.0, 0.5, 0.0}
\definecolor{ballblue}{rgb}{0.13, 0.67, 0.8}
\definecolor{darklavender}{rgb}{0.45, 0.31, 0.59}
\definecolor{chromeyellow}{rgb}{1.0, 0.65, 0.0}
\definecolor{candypink}{rgb}{0.89, 0.44, 0.48}
\definecolor{blue(ncs)}{rgb}{0.258, 0.520, 0.953}
\usepackage{color-edits}
\addauthor{gn}{magenta}
\addauthor{zw}{orange}
\addauthor{df}{cyan}
\addauthor{hz}{blue}
\addauthor{zc}{violet}

\title{What Are Tools Anyway? \\A Survey from the Language Model Perspective}
\author{Zora Zhiruo Wang$^{\spadesuit}$ \quad Zhoujun Cheng$^{\vardiamondsuit}$ \quad Hao Zhu$^{\spadesuit}$ \quad
{\bf  Daniel Fried$^{\spadesuit}$} \quad {\bf Graham Neubig}$^{\spadesuit}$ \\
$^{\spadesuit}$Carnegie Mellon University \quad $^{\vardiamondsuit}$Shanghai Jiao Tong University}

\colmfinalcopy % Uncomment for camera-ready version, but NOT for submission.
\begin{document}

\maketitle

\begin{abstract}
Language models (LMs) are powerful yet mostly for text generation tasks. Tools have substantially enhanced their performance for tasks that require complex skills.
However, many works adopt the term ``tool'' in different ways, raising the question: \textit{What is a tool anyway?} Subsequently, \textit{where and how do tools help LMs?}
In this survey, we provide a unified definition of tools as external programs used by LMs, and perform a systematic review of LM tooling scenarios and approaches.
Grounded on this review, we empirically study the efficiency of various tooling methods by measuring their required compute and performance gains on various benchmarks, and highlight some challenges and potential future research in the field.% \footnote{\url{https://github.com/zorazrw/awesome-tool-llm}}
\end{abstract}

% ############
\section{Introduction}
\label{sec:intro}

Language Models (LMs) have become increasingly effective in solving text-generation tasks, by taking in natural language (NL) instructions from users and outputting NL responses, such as answering the ``What is the capital of the US?'' with ``Washington D.C.''. However, LMs often struggle to perform tasks that require complex skills (e.g., math or complex reasoning), and are fundamentally unable to solve other tasks that require access to information not included in their training data (e.g., the current weather or date).

\begin{wrapfigure}[13]{r}{0.42\textwidth}
\vspace{-4mm}    
\includegraphics[width=0.39\textwidth]{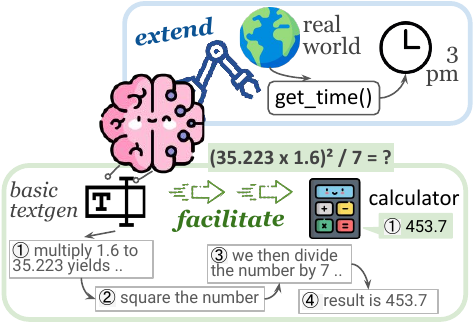}
\vspace{-2mm}
\caption{Illustration of tools extending and facilitating LM task-solving.}
\label{fig:def}
\end{wrapfigure}

To solve this problem, researchers and practitioners are turning to LMs enhanced with \emph{tools}, which help \textit{facilitate} the task-solving process of LMs, or \textit{extend} LMs with new abilities that the LM does not possess otherwise \citep{qin2023tool,mialon2023augmented}.
For example, a \texttt{calculator} tool may be used to facilitate mathematical calculations, or a \texttt{get\_time()} tool could be used to obtain the current time, which is not available purely through the LM's parameters. 
Inspired by the tools used by humans \citep{shumaker2011animal}, some works introduce application-specific \texttt{software} as tools, such as using a \texttt{search engine} to obtain knowledge \citep{lazaridou2022internetaugmented,komeili-etal-2022-internet}, using a \texttt{translator} to process unknown languages \citep{schick2023toolformer}, or using a \texttt{SQL engine} to query databases \citep{hao2023toolkengpt,zhuang2023toolqa}.
With the development of numerous application programming interfaces (APIs) on the web, many works collect \texttt{API}s as tools to access world data in real-time \citep{balog2016deepcoder,xu2023tool,qin2023toolllm} via multiple modalities \citep{tang2023toolalpaca}, even performing professional activities such as financial analysis \citep{li-etal-2023-api} and digital marketing \citep{huang2024metatool}.
Instead of using black-box APIs with unseen implementations, other works use locally-crafted \texttt{function}s to query over structured tables \citep{wang2024executable,cao2023api} or images \citep{suris2023vipergpt}, where the function tools can be created by human \citep{gupta2022visual} or model experts \citep{wang2023voyager,cai2023large,wang2024trove}.

% issue, motivation; our diff from existing surveys
However, despite this broad and burgeoning area of tool use in LMs, existing surveys only cover certain tool categories such as software \citep{mialon2023augmented} or APIs \citep{qin2023tool}. 
In this paper, we (1) provide a unified view of tool use across a broad range of scenarios, (2) empirically analyze the cost efficiency of tooling methods, to give practical guidance on when and how one should use tools, and (3) offer concrete suggestions for evaluations.

We start with proposing \textit{a unified definition} of tools and explain \textit{why tools help task-solving} (\S\ref{sec:background}).
We first introduce the \emph{basic tool-use paradigm} (\S\ref{sec:basic-paradigm}) and study a variety of tool-using scenarios by enumerating \textit{which tools exist} and \textit{to which tasks they apply} (\S\ref{sec:scenarios}). 
Next, we study advanced approaches for \textit{complex tool usage} and even \textit{make new tools} if they are unavailable for the task (\S\ref{sec:methods}). 
We then summarize existing testbeds and evaluation metrics across LM tooling works, and highlight several missing aspects with concrete metric suggestions (\S\ref{sec:good-tool}).
Lastly, grounding on our empirical analysis about \textit{when tools are effective}, we identify the most efficient tooling approaches and the tasks that benefit most from tools (\S\ref{sub:trade-off}).

% ##################
\section{Background}
\label{sec:background}

\subsection{What are tools?}
% how existing works define tools
Because LMs are products of the digital world, tools employed by LMs are often computer \textbf{programs} that are executable in corresponding environments, e.g., Python programs are executable in Python environments. 
Referring back to human-used tools, \citet{shumaker2011animal} defines animal tool use as \textit{``the external employment of an unattached or manipulable attached environmental object to alter more efficiently the form, position, or condition of another object.''}
% two properties of tools
Similar to this definition of physical tools, LM-used program tools should also be \textbf{external} to the employer (i.e., the LM) and are part of the environment. 
In the meantime, instead of arbitrary program snippets, a tool is a \textbf{function} (e.g., \texttt{plus\_one}), meaning that it can be applied to other objects (e.g., data) and yield an output (e.g. \texttt{plus\_one}$(1) \rightarrow 2$).

Existing definitions of LM-used tools touch on some of these aspects. 
\citet{qin2023tool} make an intuitive appeal to the similarity to human tool use, but do not define what entails a tool.
\citet{mialon2023augmented} define \textit{a tool} as \textit{``an external module that is typically called using a rule or a special token and whose output is included in the augmented LM's context.''}
We argue for a somewhat more broad definition than this, which encompasses a wide variety of more recent works on tool usage:

\begin{definition}
  \label{def:tools-lm}
  An LM-used tool is a function interface to a computer program that runs \textit{externally} to the LM, where the LM generates the function calls and input arguments in order to use the tool.\footnote{Under our definition, tool functions can be implemented by any means, including symbolic computation or neural networks --- the functions only require a programmatic interface.}
\end{definition}

\subsection{Why are tools helpful?}
Tools can help task-solving in different ways, depending on the functionality of individual tools. We summarize their functions into three major categories: perception, action, and computation. A tool may belong to one or more of these three categories.

\noindent \textbf{Perception} \quad
Perception tools provide or collect information from the environment. An example is using a \texttt{get\_time()} API to obtain the current time, which is not included in the LM's parametric knowledge learned from training.

\noindent \textbf{Action} \quad
Action tools can exert actions on the environment and change its state. For example, \texttt{turn\_left()} can shift the direction of an embodied agent, or executing \texttt{make\_post(website, post)} can change the content on a \texttt{website}.

\noindent \textbf{Computation} \quad
Computation tools do not necessarily perceive or modify the external environment, but use programs to tackle complex computational tasks. For example, a \texttt{calculator} is a computation tool for mathematical calculation.
Note that the computation also includes more general acts of computing beyond numerical calculation. Therefore, a \texttt{translator} is also a computation tool that can be used to translate between languages.

Note that many tools can fall into multiple categories.
For instance, a \texttt{search engine} is a tool that can perform both computation and perception.
As computation, it measures document similarity and selects relevant ones, but it also perceives the environment (i.e., the web) and fetches data (i.e., returned documents) from it.
In a similar spirit, \textsc{SQL} queries can be used as computation tools (e.g., \texttt{SELECT SQRT(16) / 10 AS result}), perception tools for viewing data (e.g., \texttt{SELECT name FROM data}), action tools to modify data (e.g., \texttt{INSERT INTO data VALUES name}), or all of the above (e.g., \texttt{INSERT INTO counts (grp\_id, grp\_cnt) SELECT grp\_id, COUNT(*) FROM data GROUP BY grp\_id}).

\subsection{Tools and ``Agents''} 
There has recently been a burgeoning of work on LM-powered agents \citep{xi2023rise,sumers2024cognitive}.
\citet{russell2010artificial} define agents as \textit{``anything that can be viewed as perceiving its environment through sensors and acting upon that environment through actuators.''}
According to this definition, agents are programs that use perception tools to perceive the situated environment, or action tools to interact with the environment.
Models that only use computation tools and do not interact with their environments through perception or action tools arguably do not fall under the category of ``agents'' according to this definition.

% ###################################
\section{The basic tool use paradigm}
\label{sec:basic-paradigm}
% figure start
\begin{wrapfigure}[16]{r}{0.42\textwidth}
\vspace{-5mm}
\includegraphics[width=0.40\textwidth]{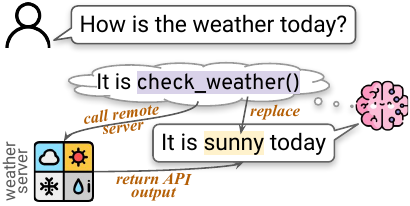}
\vspace{-2mm}
\caption{The basic tool use paradigm. LM calls \raisebox{0.3mm}{\colorbox{violet!13}{\texttt{check\_weather}}} tool by generating text tokens. This call triggers the server to execute the call and return the output \raisebox{0.3mm}{\colorbox{yellow!33}{sunny}}, using which the LM replaces the API call tokens in the response to the user.}
\label{fig:call-api}
\end{wrapfigure}
% figure end

First, in this section, we show an illustrative example of a basic tool-use paradigm introduced by Toolformer \citep{schick2023toolformer}, which many tool-related works adopt (\autoref{fig:call-api}). Assuming an LM communicates with users mainly in natural language, upon receiving a user query such as ``How is the weather today?'', the LM then proceeds to generate either text or tool calls. In the example, starts with generating a few tokens of text ``It is ...''. When the LM needs to seek external tools to complete the task, e.g., get real-time weather information, it generates tokens of the tool name and corresponding input arguments enclosed with \texttt{(}parentheses\texttt{)} to construct a complete tool calling expression. 
This completed expression will trigger a shift from text-generation mode to tool-execution mode. The server hosting the tool will execute the expression and return the execution result to the LM. 

Taking the example in \autoref{fig:call-api}, the LM sends the \texttt{check\_weather()} call to the weather server and receives the output ``sunny''. The returned result replaces the tool call in the LM-generated tokens (e.g., from ``It is \texttt{check\_weather()}'' to ``It is sunny''), which is used for subsequent steps of generation.
Accordingly, the LM shifts back to the text generation mode and continues to finish the response by generating new text tokens, e.g., adding `today.', and finally returning the response to the user.

In order for LMs to use this basic paradigm of using tools, current works mainly leverage inference-time prompting and training-time learning methods.

\noindent \textbf{Inference-time prompting} \quad
Leveraging the ability of LMs to learn in-context \citep{brown2020language}, many works provide tool information through a prompt and expect LMs to acquire abilities to use these tools from input contexts.
This is achieved by providing instructions about the task, example pairs of queries and solutions that use tools \citep{gupta2022visual,lu2023chameleon,paranjape2023art,shen2023hugginggpt,yang2023mmreact}, and/or documentation of the tools' functionality \citep{hsieh2023tool}.

\noindent \textbf{Learning by training} \quad
Beyond learning tools from test-time contexts, LMs can learn from examples that use these tools during training.
LMs can simply be trained to generate tool-using solutions, where the examples can be manually annotated by humans \citep{li-etal-2023-api}, synthesized by larger teacher LMs \citep{tang2023toolalpaca,qin2023toolllm,huang2024metatool}, or bootstrapped by the test-time LM itself \citep{schick2023toolformer}.

\textbf{\begin{table}[t]
\vspace{-5mm}
\small
\begin{center}
    \begin{tabular}{ll}
    \toprule
    \multicolumn{1}{c}{\textbf{Category}} & \multicolumn{1}{c}{\textbf{Example Tools}} \\
    \midrule
    \multirow{3}{*}{\kn ~~Knowledge access} & {\texttt{sql\_executor(query:~str) -> answer:~any}} \\
    {} & {\texttt{search\_engine(query:~str) -> document:~str}} \\
    {} & {\texttt{retriever(query:~str) -> document:~str}} \\
    \midrule
    \multirow{3}{*}{\work ~~Computation activities} & {\texttt{calculator(formula:~str) -> value:~int | float}} \\
    {} & {\texttt{python\_interpreter(program:~str) -> result:~any}} \\
    {} & {\texttt{worksheet.insert\_row(row:~list, index:~int) -> None}} \\
    \midrule
    \multirow{4}{*}{\world ~~Interaction w/ the world} & \texttt{get\_weather(city\_name:~str) -> weather:~str} \\
    {} & \texttt{get\_location(ip:~str) -> location:~str} \\
    {} & \texttt{calendar.fetch\_events(date:~str) -> events:~list} \\
    {} & \texttt{email.verify(address:~str) -> result:~bool} \\
    \midrule
    \multirow{3}{*}{\modal ~~Non-textual modalities} & \texttt{cat\_image.delete(image\_id:~str) -> None} \\
    {} & \texttt{spotify.play\_music(name:~str) -> None} \\
    {} & \texttt{visual\_qa(query:~str, image:~Image) -> answer:~str} \\
    \midrule
    \multirow{2}{*}{\nn ~~Special-skilled LMs} & {\texttt{QA(question:~str) -> answer:~str}} \\
    {} & \texttt{translation(text:~str, language:~str) -> text:~str} \\
    \bottomrule
    \end{tabular}
\end{center}
\vspace{-2mm}
\caption{Exemplar tools for each category.}
\vspace{-15mm}
\label{tab:example-tools}
\end{table}}

\section{Scenarios where tools are useful}
\label{sec:scenarios}

While LMs may easily learn to do many tasks to high accuracy without tools, many other tasks greatly benefit from tool use.
In this section, we study a broad range of scenarios where tools have been used to assist agents. We discuss tasks where human-created, application-specific tools can improve their performance or other positive aspects (\S\ref{sub:app-spec}), as well as scenarios where tools may not be as useful (\S\ref{sub:not-useful}).

\subsection{Utilizing existing tools for specific applications}
\label{sub:app-spec}

While it is difficult to exhaustively enumerate every scenario where tools could be useful, we summarize some major categories of tool use in \autoref{tab:example-tools} and below.
Note that a tool may fall into one or more categories.

\noindent \textbf{\raisebox{-1mm}{\includegraphics[width=4mm]{./figures/icon/book-icon.png}} Knowledge access} \quad
% unstructured text & structured kb/kg
LMs store limited knowledge during training due to both limits in (i) the data that they are trained on and (ii) the ability of LMs to accurately memorize and utilize all of the data that they see at training time.
Several varieties of tools can be used to alleviate this issue.
\texttt{SQL} and \texttt{SPARL} executors can provide access to data in structured knowledge bases \citep{thoppilan2022lamda,parisi2022talm,hao2023toolkengpt} or knowledge graphs \citep{zhuang2023toolqa}.
An \texttt{search engine} tool  over the Internet \citep{yao2023react,schick2023toolformer,paranjape2023art} can enable LMs to access more up-to-date information \citep{komeili-etal-2022-internet,lazaridou2022internetaugmented}.
More generally, retrieval-augmented generation systems \citep{asai2023retrieval} can be seen as using a \texttt{retriever} tool \citep{mialon2023augmented}.

\noindent \textbf{\raisebox{-0.5mm}{\includegraphics[width=4mm]{./figures/icon/work-icon.png}} Computation activities} \quad
% math
Complex computing activities such as math calculations are known to be challenging for neural LMs \citep{schick2023toolformer}. While even a \texttt{calculator} can enhance LMs' numeracy abilities \citep{parisi2022talm,hao2023toolkengpt}, more generic \texttt{Python} programs are also employed to aid reasoning tasks \citep{gao2023pal,chen2023program,wang2023mint}.
% business tools
For more complex professional jobs, business tools are also applied, such as using \texttt{worksheet} to manipulate Google Sheets \citep{xu2023tool}, or even tools for financial, medical, education, or advertising domains \citep{tang2023toolalpaca,huang2024metatool}.

\noindent \textbf{\raisebox{-0.5mm}{\includegraphics[width=4mm]{./figures/icon/world-icon.png}} Interaction with the world} \quad
LMs without tools are fundamentally unable to perceive and act in the world around them, necessitating tool use where such perception and action is necessary. For instance, LMs can access real-time information such as weather \citep{xu2023tool,tang2023toolalpaca}, or positional knowledge such as location \citep{qin2023toolllm}. On the other hand, LMs can manipulate real-world information such as managing calendars \citep{schick2023toolformer} and emails \citep{qin2023toolllm}.
In addition to web-based activities, LMs can engage in physical activities in embodied environments, such as fishing with rods or mining with axes in the Minecraft world \citep{wang2023voyager}; further propagate to the real-world tasks to perform cooking \citep{singh2022progprompt,shridhar2020alfred}, plotting \citep{liang2023code}, and even conducting chemical research \citep{boiko2023autonomous}.

\noindent \textbf{\raisebox{-0.5mm}{\includegraphics[width=4mm]{./figures/icon/media-icon.png}} Non-textual modalities} \quad
While many LMs only consume and generate texts, some works bring in access to visual \citep{gupta2022visual,suris2023vipergpt}, audio \citep{yang2023mmreact,gao2023assistgpt}, or other modalities. For example, LMs can access images with \texttt{cat\_image} APIs \citep{xu2023tool,tang2023toolalpaca} or songs \citep{huang2024metatool} provided by \texttt{spotify}, even answer questions about them \citep{gupta2022visual,gao2023assistgpt}.

\noindent \textbf{\raisebox{-0.5mm}{\includegraphics[width=4mm]{./figures/icon/brain-icon.png}} Accessing specialized LMs} \quad
Some works propose to use specialized LMs as tools, essentially using the main LM as a task planner to dispatch requests to other LMs.
\citet{schick2023toolformer} propose QA models to fill in factoid details in responses, \citet{thoppilan2022lamda,schick2023toolformer,paranjape2023art} use machine translation models to assist multilingual tasks. Beyond specific tasks, some works adopt multiple neural models from Hugginface or similar platforms \citep{patil2023gorilla,shen2023hugginggpt}, or further fine-tune them on various data \citep{viswanathan2023prompt2model}.
Compared to the base LM, these tool models mainly vary in their specialized skills, and may or may not have substantial architectural differences from the base LMs.

\subsection{Where are tools \emph{not} useful?}
\label{sub:not-useful}

Despite the fact that tools can be helpful under many scenarios discussed above, it is also important to note scenarios where tools are arguably not very helpful.
Some examples of tasks where tools have not (yet) been used to great effect include machine translation, summarization, and sentiment analysis (among others).
These are tasks that are not easy to perform using non-ML methods (c.f.~solving math problems or accessing databases, which can be done using a calculator or SQL), and can be performed with high accuracy by a powerful LM alone.
One intuitive reason is that the tools currently leveraged for these tasks are \textit{neural networks} and have limited advantages over the base LM.
Imagine if we leverage tools on these tasks, the tools would mostly generally be another neural LM with specialized skills, e.g., an LM specifically trained on many summarization datasets to perform this task. However, this special-skilled neural LM may not have significant architectural differences from the base tool-using LM, or be smaller in size or training tokens hence having inferior language modeling abilities in general.
In comparison, the base LM capable of solution planning and tool management, usually are more powerful (e.g., GPT-4) and can achieve reasonable performance on a wide variety of tasks, perhaps even outperforming special-purpose LMs \citep{robinson-etal-2023-chatgpt}.

\section{Advanced tool-use methods}
\label{sec:methods}

Given this understanding of the basic tooling paradigm and the scenarios in which tools are useful, we now discuss more advanced approaches for tools.
Concretely, we study multi-tool selection and usage (\S\ref{sub:tool-select}), complex tooling under programmatic contexts (\S\ref{sub:tools-and-programs}), and creation of tools when they are not available a-priori (\S\ref{sub:make-tool}).

% #########################
\subsection{Complex tool selection and usage}
\label{sub:tool-select}

Depending on the number of tools available, the system may include an implicit or explicit tool selection module.
If tools are already \textit{designated} for the task \citep{lazaridou2022internetaugmented,thoppilan2022lamda}, then no tool selection is needed.
If \textit{a small number} (e.g., 5--10) of tools are available, metadata and use cases of these tools can be provided as input contexts along with the user query \citep{schick2023toolformer,paranjape2023art}, and LMs can directly select tools from contexts via a standard generation process.
If the toolbox size \textit{further grows} (e.g., to hundreds), fitting all tools into model inputs is not feasible. Thus an extra retrieval step is often incorporated: a retriever model short-lists the most relevant tools and feeds their metadata to the solution-generation LM. Specifically, \citet{zhou2023docprompting,qin2023toolllm} train retriever models that map NL intents to tool documentation.
\citet{yuan2023craft} ask LMs to write hypothetical descriptions and use the SimCSE retriever \citep{gao2021simcse} to find similar tools. More easily, one can directly use off-the-shelf embeddings \citep{SFRAIResearch2024,openai-emb} or training-free sparse retrievers \citep{robertson2009probabilistic}.

% \noindent \textbf{Complex, multi-tool usage} \quad
For complex queries that require multiple tools to solve, the common approach so far is to break down the task and tackle each step sequentially \citep{paranjape2023art} by selecting and using tools with intermediate contexts.
However, this sequential multi-turn paradigm may not be reflective of more complex or realistic usage of the involved tools. For example, a user may prefer \textit{nested} function calls \texttt{check\_weather(get\_local\_time(`Pittsburgh'))} to allow information hiding or encapsulation \citep{rogers2001encapsulation}, \textit{parallel} calls to reduce round trips with the API \citep{eleti2023function}, or \textit{iterative} calls \texttt{buy\_ticket(event)} in a loop until it returns \texttt{True} to indicate a successful transaction.

% ################################
\subsection{Tools in programmatic contexts}
\label{sub:tools-and-programs}

Unlike text-based tasks where tools are auxiliary modules to extend LM abilities, on programmatic tasks, where code LMs can solve the problem by generating programs, tools can be seen as compositions of basic functions.
In this part, we discuss tools in programmatic tasks for domain-specific (\S\ref{sub:domain-spec}) and general-purpose problems (\S\ref{sub:general-codegen}).

\noindent \textbf{Focus on varied tools} \quad
Depending on the tasks of interest, existing works focus on different types of tools under programmatic contexts. With the increasing complexity of these tools and presumably a decreasing familiarity of LMs about them, there are works that adopt (i) \textcolor{candypink}{\textit{built-in functions}} of a programming language (PL) to augment LMs in symbolic reasoning, (ii) \textcolor{ao!80}{\textit{external libraries}} in pre-designed packages to tackle complex open-domain coding queries \citep{wang2023execution}, and (iii) \textcolor{blue(ncs)}{\textit{utility functions}} unseen at training time to solve specific tasks.

\begin{figure}[ht]
\centering
\vspace{-1mm}
    \includegraphics[width=\textwidth]{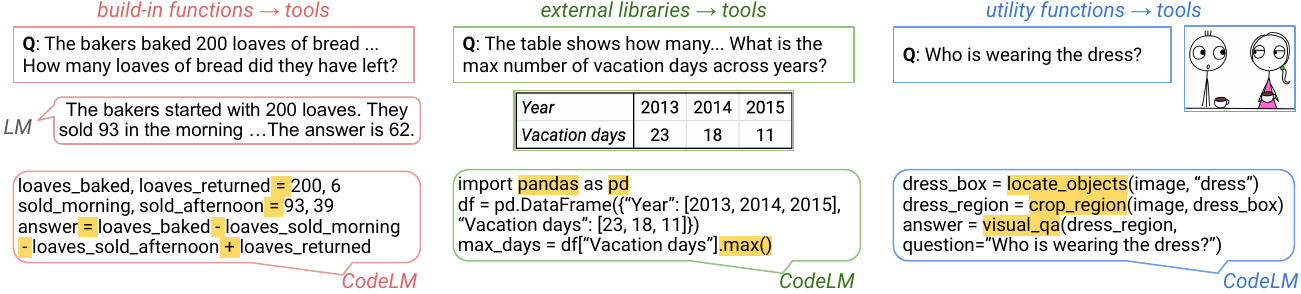}
\vspace{-6mm}
\caption{Relative to what is considered as the base LM or base actions, tools can refer to built-in functions, external libraries, or task-specific utility functions (from left to right).}
\vspace{-1mm}
\label{fig:codelm-tools}
\end{figure}

% ############
\subsubsection{Domain-specific semantic parsing}
\label{sub:domain-spec}

NL-to-code generation systems have been studied for many years on special-domain tasks such as querying databases \citep{zelle1996learning,zettlemoyer2012learning} or knowledge graphs \citep{berant2013semantic}. Code produced by these systems is often domain-specific logical forms (DSL) manually designed by experts, such as lambda expressions \citep{liang2013lambda} or SQL queries \citep{yu2018spider}, and more recently, the QDMR grammar \citep{wolfson2020break} as an extension to SQL.
In addition to knowledge-oriented tasks, many agentic tasks adopt DSL to operate in corresponding environments, such as \texttt{click} or \texttt{type} in web navigation \citep{zheran2018reinforcement,webshop2022yao,zhou2024webarena}, \texttt{placeItem} and \texttt{killMob} in the embodied Minecraft world \citep{wang2023voyager}, or \texttt{set\_joint\_target} for robot dogs \citep{yu2023language}.
Because DSLs are often specific enough to the target problems, most works directly use these \textcolor{candypink}{built-in actions}.
Yet still, for complex task queries, solution programs written in basic DSL actions alone can be hard to interpret or cumbersome to use, e.g., it is hard to tell that the lambda expression \texttt{(fold xs ($\lambda$ (n x) (+ 1 n)) 0)} is to calculate the length of \texttt{xs}.

% ############
\subsubsection{General-purpose code generation}
\label{sub:general-codegen}
Recent code generation systems have expanded from using DSL to more general-purpose PLs such as Python or Java \citep{yin-neubig-2017-syntactic,chen2021evaluating}. These languages enable more programming flexibility and readily apply to versatile scenarios. 
As we have introduced using \textcolor{candypink}{built-in actions} as tools in \S\ref{sub:app-spec}, we discuss more on two other
common categories of tools for code LMs, namely \textcolor{ao!90}{\textit{external libraries}} and task-specific \textcolor{blue(ncs)}{\textit{utility functions}}.

\noindent \textbf{External libraries} \quad
From the usage of PLs, built-in functions are internal to whichever environment, whereas third-party libraries lie externally and need to be imported to tackle specific contexts, such as \autoref{fig:codelm-tools} (middle). Aligning with this conception, \citet{zhang2023toolcoder} use Python libraries such as \texttt{matplotlib} to plot figures and \texttt{pandas} to manage data.

\noindent \textbf{Utility functions} \quad
For more task-specific applications,
expert-crafted utility functions, usually unseen at training time, are incorporated as tools. E.g., in \autoref{fig:codelm-tools} (right), the highlighted \texttt{locate\_objects} function is designed by human experts \citep{gupta2022visual,suris2023vipergpt} to load neural models and perform post-processing to obtain the detected box region. In a similar spirit, \citet{cheng2023binding} use GPT as a tool to query world facts external to the tabular contents, \citet{cao2023api} further design macro operation APIs to support advanced tabular operations.
However, because human tool curation requires expertise and effort, some works explore using LMs to automatically create tools instead.

% ################################

\begin{wrapfigure}[8]{r}{0.40\textwidth}
\vspace{-6mm}
\includegraphics[width=0.39\textwidth]{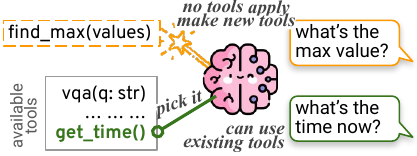}
\vspace{-2mm}
\caption{LM makes tools when no tools readily apply to the task.}
% \vspace{-1mm}
\label{fig:make-tool}
\end{wrapfigure}

\subsection{Tool creation and reuse}
\label{sub:make-tool}

While one can readily use tools for tasks equipped with pre-designed tools, for tasks that do not have readily-applicable, human-created tools, some works explore using LMs to make tools and use them.

\noindent \textbf{Domain-specific library abstraction} \quad
Works that use DSLs often compose frequently-used-together actions as shortcut tools. For example, \citet{ellis2023dreamcoder} learn function abstractions such as \texttt{length} and \texttt{count\_to} from lambda primitives (e.g., \texttt{$0$}, \texttt{+}) for the list processing task. 
\citet{pmlr-v139-wong21a,bowers2023top} similarly build functions bottom-up from a large corpus of DSL programs. More recently, \citet{grand2023lilo} use LLMs to abstract libraries with auto-documentation. 
Further for agentic tasks, \citet{zheran2018reinforcement} learn common workflows to guide web navigation, such as composing the basic $\{$\texttt{click}, \texttt{like}$\}$ actions to form a higher-level login action \texttt{click(like(`login'))}.

\noindent \textbf{General-purpose tool making} \quad
Nonetheless, on general-purpose PLs, running the DSL-oriented methods above may expand their search space and limit their scalability. Instead, recent works often leverage LMs' procedural knowledge to alleviate the search issue. 
To start, \citet{wang2023voyager} designs an automatic learning curriculum in Minecraft to make and use Java program tools.
LATM \citep{cai2023large} use LMs to build, verify, and use Python tools on BigBench \citep{srivastava2023beyond} tasks, where however, all examples require the same single tool hence have limited difficulty.
CREATOR \citep{qian2023creator} extend tool-making to harder tasks such as math and table world problems, and improves task success by creating tools yet repetitively for individual examples, thus CRAFT \citep{yuan2023craft} add heuristic-based training to craft less repetitive tools.
Towards more efficient pipelines, ReGAL \citep{stengeleskin2024regal} learns from refactoring a smaller number of programs, while TroVE \citep{wang2024trove} purely relies on inference-time execution signal and induces reusable tools on-the-fly.

\section{How to evaluate tool use?}
\label{sec:good-tool}

In this section, we study existing LM-tooling benchmarks (\S\ref{sub:testbeds}) and their evaluation metrics (\S\ref{sub:eval-metrics}), lastly, we discuss the missing yet important evaluation aspects of tools (\S\ref{sub:desired-properties}).

% ############
\subsection{Testbeds for evaluating tools}
\label{sub:testbeds}
LM tool use can be evaluated on (i) repurposed existing datasets that can additionally benefit from tools (\S\ref{sub:existing-dataset}), and (ii) newly crafted benchmarks that necessitate tool use (\S\ref{sub:api-benchmark}).

% ############
\subsubsection{Repurposed Existing Datasets}
\label{sub:existing-dataset}
Many tasks are solvable by using LMs, yet often with great difficulty or inefficiency. Therefore, some works use tool-augmented LMs as an alternative approach to solve these tasks. 

Many of these datasets require \textbf{reasoning}. Starting from when questions are expressed in NL, such as complex reasoning with the Big-bench \citep{srivastava2023beyond} dataset, mathematical problems with the MATH \citep{hendrycks2021measuring} dataset, and reasoning over world knowledge to answer questions in NaturalQuestions \citep{kwiatkowski2019natural} and TriviaQA \citep{joshi2017triviaqa} datasets.
Beyond free-form texts, datasets that require reasoning over \textbf{structured data} can also benefit from tools. These tasks include table-based QA with tabular math world problems in TabMWP \citep{lu2023dynamic}, Wikipedia tables in WTQ \citep{pasupat-liang-2015-compositional}, and complex-structured tables in HiTab \citep{cheng-etal-2022-hitab}.
Beyond the text modality, datasets that require reasoning over \textbf{other modalities} also benefit from modality-extending tools, e.g., answering questions about an image with the GQA \citep{hudson2019gqa} dataset, or image pairs with the NLVR2 dataset \citep{suhr2019corpus}.

Because tool use is proposed as an alternative method to solve these datasets, evaluations of these tool-augmented systems follow the standard evaluation process for individual datasets. Concretely, almost all tasks are measured by answer exact match, either in textual or numerical formats.
Note that, to obtain the final answers for lexical matching evaluations, all tool-calling expressions \textbf{need to be executed}, and the execution outputs are incorporated into the final answers produced by the tool-augmented systems, as introduced in \S\ref{sec:basic-paradigm}.

% ############
\subsubsection{Aggregated API Benchmarks}
\label{sub:api-benchmark}
Existing benchmarks can only benefit from a limited set of tools, yet there are far more tools we can utilize to perform versatile tasks in the real world, particularly the API tools created by human developers spread on the web. 
Therefore, many recent works aggregate API tools from various web sources and create benchmarks for using these APIs, as shown in \autoref{tab:api-benchmarks}.

\begin{table}[ht]
\small
\vspace{-1mm}
\begin{center}
    \begin{tabular}{l|lllc}
    \toprule
    \multicolumn{1}{c|}{\textbf{Benchmark}} & \multicolumn{1}{c}{\textbf{Tool Source}} & \multicolumn{1}{c}{\textbf{Example Curation}} & \textbf{Domain (\S\ref{sub:app-spec})} & \multicolumn{1}{c}{\textbf{Executable}} \\ 
    \midrule
    {\hyperlink{cite.xu2023tool}{ToolBench$_1$}} & {existing dataset} & {adopted, human annotated} & {\work, \world} & {\ding{51}} \\
    {\hyperlink{cite.qin2023toolllm}{ToolBench$_2$}} & {RapidAPI} & {model synthesized} & {\work, \world} & {\ding{51}} \\
    {\hyperlink{cite.zhuang2023toolqa}{ToolQA}} & {existing dataset} & {model synthesized} & {\work, \kn} & {\ding{51}} \\
    {\hyperlink{cite.tang2023toolalpaca}{ToolAlpaca}} & {PublicAPIs} & {model synthesized} & {\kn, \work, \world, \modal} & {\ding{55}} \\
    {\hyperlink{cite.li-etal-2023-api}{API-Bank}} & {PublicAPIs} & {human annotated} & {\work, \world} & {\ding{51}} \\
    {\hyperlink{cite.huang2024metatool}{MetaTool}} & {OpenAI Plugins} & {model synthesized} & {\work, \world, \modal} & {\ding{55}} \\
    % \midrule
    {\hyperlink{cite.patil2023gorilla}{Gorilla}} & {HF, Torch, TF} & {model synthesized} & {\nn} & {\ding{55}} \\
    {\hyperlink{cite.shen2023hugginggpt}{HuggingGPT}} & {HF} & {human annotated} & {\nn} & {$~~$\ding{55}$^{*}$}\\
    {\hyperlink{cite.shen2023taskbench}{Task Bench}} & {HF, PublicAPIs} & {model synthesized} & {\nn, \modal, \world} & {\ding{55}} \\
    \bottomrule
    \end{tabular}
\end{center}
\vspace{-3mm}
\caption{Benchmarks of providing aggregated APIs to LMs as tools. 
HF is short for HuggingFace. `\ding{55}$^{*}$' means that: though tools employed by HuggingGPT are executable, it does not evaluate the execution output due to the cost of hosting and inferencing.}
\vspace{-1mm}
\label{tab:api-benchmarks}
\end{table}

\noindent \textbf{Tool sources} \quad
Tools are mainly aggregated from existing datasets or public APIs.
While \citet{xu2023tool,zhuang2023toolqa} adopt existing datasets and propose alternative methods via tool augmentation, these benchmarks are limited in domains. Several works scrape more APIs from online sources such as Public APIs \citep{tang2023toolalpaca}, RESTful APIs \citep{tang2023toolalpaca}, or the OpenAI plugin list \citep{huang2024metatool}. Beyond human-crafted APIs \citep{li-etal-2023-api}, neural models from ML platforms can be similarly presented in an API format \citep{patil2023gorilla,shen2023hugginggpt,shen2023taskbench}. 
Nonetheless, as tools are collected from heterogeneous sources, it is challenging to select the best benchmark or unify all these varied benchmarks.

\noindent \textbf{Example curation} \quad
Examples can be adopted from existing datasets, annotated by humans, or synthesized by LMs.
While most examples adopted from existing datasets are human annotated \citep{xu2023tool}, only \citet{li-etal-2023-api} do so for scraped APIs, by surveying 500 people and creating 314 dialogues manually.
Most other works prompt GPT models to synthesize examples \citep{qin2023toolllm,tang2023toolalpaca,shen2023taskbench,zhuang2023toolqa, huang2024metatool}, however, leading to issues of \textit{naturalness} and \textit{executability}.

\textbf{First}, LMs are often asked to create examples, even tool outputs in \citet{tang2023toolalpaca}, given a heuristically selected set of tools. This approach leads to potential issues in two-fold: (i) the selected tools may not be used together in practice, and (ii) the synthesized examples may not reflect the \textit{natural use cases} of these tools.
\textbf{Second}, 5 out of 9 benchmarks in \autoref{tab:api-benchmarks} do not support tool execution, to alleviate the cost of hosting multiple APIs, especially when they may fail or produce unstable outputs. For example, the weather returned by the \texttt{check\_weather} API may change over time. This un-executability causes \textit{issues in evaluation}. Instead of matching final execution results using lexical- \citep{li-etal-2023-api} or neural-based metrics \citep{tang2023toolalpaca, qin2023toolllm}, works with unexecutable tools resort to pseudo matching of API calling expressions with lexical~\citep{tang2023toolalpaca,shen2023hugginggpt,huang2024metatool} and syntactical \citep{patil2023gorilla, shen2023taskbench} means.

% #################################
\subsection{What metrics are measured now?}
\label{sub:eval-metrics}

\noindent \textbf{Task completion} \quad
Tools are used to assist task solving. Most works that allow tool execution evaluate the task completion score to quantify the effectiveness of utilizing tools.

\noindent \textbf{Tool selection} \quad
For datasets with execution issues \citep{huang2024metatool,shen2023taskbench}, another common metric is the accuracy of selecting the correct tools. This helps disentangle incorrect tool selection errors from inaccurate tool usage errors. 
Despite that tool selection mainly serves as a proxy for evaluating task completion when having unexecutable tools, it can be seen as a measure of LM planning abilities --- the process of breaking down a task into multiple steps and selecting tools to complete individual steps.

\noindent \textbf{Tool reusability} \quad
While tool reusability is often deemed important in took-making literature \citep{cai2023large,yuan2023craft}, only \citet{wang2024trove} evaluates tool reusability by the size of induced toolboxes over a fixed number of examples. As its literal meaning, reusable tools can be (re)used to solve multiple examples hence having more generic functionalities. Adopting a reusable tool is more efficient than using multiple specific tools, and facilitates human verification in both speed and accuracy dimensions \citep{wang2024trove}.

% #################################
\subsection{What properties are missing?}
\label{sub:desired-properties}

\noindent \textbf{Efficiency of tool integration} \quad
As demonstrated by our empirical study (\S\ref{sub:trade-off}), the benefits brought by the tools come with the cost of additional computation, especially for teaching LMs to use tools via training or prompting. In addition to performance gain, reporting the computation overhead can enable fairer comparisons between different approaches.

\noindent \textbf{Quality of tools} \quad
While existing works mostly focus on how tools improve task accuracy, the \textit{performance of tools} themselves is also important. Tool performance can cover multiple aspects such as completing the call quickly, requiring less computation, and not putting users at risk or failing unexpectedly. 
One way to measure these aspects is to conduct API testing \citep{yasar2022software,ehsan2022restful} on their runtime, memory usage, and success rate.

\noindent \textbf{Reliability of unstable tools} \quad
Particularly for tools that involve \textit{neural models} or \textit{randomized components}, their output quality may be unstable and unpredictable. For example, the \texttt{VQA} tool \citep{gupta2022visual} may answer some questions correctly but others incorrectly.
It is important to \textit{be aware of} this uncertainty in contrast to stable, rule-based tools such as a \texttt{calculator}, further alleviate this instability and guarantee more predictable outputs.

\noindent \textbf{Reproducible testing} \quad
Many tools interact with the real world and may return different results at different times. For example, \texttt{check\_weather} may return ``sunny'' today but ``cloudy'' tomorrow. This irreproducible behavior poses great challenges to creating \textit{static evaluation} benchmarks with reference answers. % For instance, the answer to ``How's the weather today?'' should not be a fixed ``sunny'', because the correct answer may change according to the specific time of evaluation.
While some works alleviate this by evaluating API calls without executing them, a more rigorous method could be \textit{parallel testing} \citep{sharma2018automated} --- executing the model-generated program and the reference program in parallel, and measuring if their final outputs match.

\noindent \textbf{Safe usage} \quad
Most systems may only opt to use tools if they are trusted to be secure \citep{barbir2007challenges}.
At the very least, users favor tools that can be easily understood and verified. Further, systems may need to enforce mutual authentication and ensure data integrity \citep{ehsan2022restful}.
Yet there are more security threats and methods beyond the discussion here. We encourage readers to peruse the referenced works above for thorough studies.

\section{Trade-offs in tool usage}
\label{sub:trade-off}

Leveraging tools often brings better performance, however, should we always use tools? More concretely, is the performance gain from using tools worthy of the computation cost spent for LMs to learn to use tools, or the inference cost at test time?
Existing works mainly focus on task accuracy, but a more nuanced picture emerges when we take other factors into account.
We empirically study the performance gain and learning cost of various methods on their experimented datasets in \autoref{tab:compute-cost}, using which we discover more efficient (i.e., achieve greater gains with less compute) methods and tasks that benefit more from tools.

\begin{table}[ht]
\vspace{-3mm}
\small
\begin{center}
\resizebox{0.92\linewidth}{!}{
    \begin{tabular}{llcrcrr}
    \toprule
    \multicolumn{1}{c}{\multirow{2}{*}{\textbf{Type}}} & \multirow{2}{*}{\textbf{Method}} & \multicolumn{1}{c}{\multirow{2}{*}{\textbf{Task}}} & \multicolumn{1}{c}{\multirow{2}{*}{\textbf{$\Delta$ Perf.}}} & \multirow{2}{*}{\textbf{\# Params (B)}} & \multicolumn{2}{c}{\textbf{\# Tokens (M)}} \\
    {} & {} & {} & {} & {} & {train} & {test} \\
    \midrule
    \multirowcell{9}{tool\\use} & \multirow{5}{*}{ToolFormer} & \color{blue} {cloze} & {+ 14.7} & {6.7} & {642.1} & {269.0} \\
    {} & {} & \color{ao} {math} & {+ 30.4} & {6.7} & {3864.2} & {421.0} \\
    {} & {} & \color{chromeyellow} {QA} & {+ 5.8} & {6.7} & {1101.2} & {189.0} \\
    {} & {} & \color{red} {multilingual} & \colorbox{red!27}{- 0.2}  & {6.7} & {606.0} & {274.0} \\
    {} & {} & \color{ballblue} {temporal} & {+ 13.0} & {6.7} & {508.8} & {202.0} \\
    \cmidrule{2-7}
    {} & {API-Bank} & {API} & {+ 24.4} & {7} & \textbf{190414.6} & {0.0} \\
    \cmidrule{2-7}
    {} & {ToolAlpaca} & {API} & {+ 45.2} & {7} & \textbf{241889.3} & {0.0} \\
    \cmidrule{2-7}
    {} & \multirow{2}{*}{Chameleon} & {science} & {+ 2.6} & {-} & {0.0} & {88.3} \\
    {} & {} & \color{darklavender} {table} & {+ 1.9} & {-} & {0.0} & {325.9} \\
    \midrule
    \midrule
    \multirowcell{7}{tool\\making} & {LATM} & {BigBench} & {+ 29.1} & {-} & {28.5} & {4720.0} \\
    \cmidrule{2-7}
    {} & \multirow{2}{*}{CREATOR} & \color{ao} {math} & {+ 4.5} & {-} & {0.0} & {5113.6} \\
    {} & {} & \color{darklavender} {table} & \colorbox{chromeyellow!20}{+ 0.0} & {-} & {0.0} & \textbf{6827.6} \\
    \cmidrule{2-7}
    {} & \multirow{2}{*}{CRAFT} & \color{ao} {math} & {+ 13.2} & {-} & {4126.6} & {4098.5} \\
    {} & {} & \color{darklavender} {table} & {+ 17.2} & {-} & {2750.6} & {5018.2} \\
    \cmidrule{2-7}
    {} & \multirow{2}{*}{TroVE} & \color{ao} {math} & {+ 21.0} & {-} & {0.0} & {1825.2} \\
    {} & {} & \color{darklavender} {table} & {+ 12.0} & {-} & {0.0} & {1358.8} \\
    \bottomrule
    \end{tabular}
    }
\end{center}
\vspace{-2mm}
\caption{Computation cost (number of tokens in $M$ and parameters in $B$) of tooling methods and their performance gain on experimented datasets. To fairly compare costs on datasets with different sizes, we report the average number of tokens spent on a testing example.}
\vspace{-3mm}
\label{tab:compute-cost}
\end{table}

For each work and each dataset they experimented with,\footnote{ We did not measure some works due to insufficient resources.} we evaluate the performance gain after LM learned or made tools to solve tasks, compared to the baseline LM with no prior exposure to tool-related information.
We also quantify the computation cost of their tooling approaches during the token-consuming training and inference processes. For works using models with known sizes, we report both (i) the number of tokens in input prompts and outputs, and (ii) the parameters in experimented models to achieve corresponding performance improvements. For methods using the size-unknown GPT-4 model, which are also comparable w.r.t. to model size since they use the same GPT-4 model, we only report the number of tokens processed.
We elaborate more on computation details in \S\ref{app:learning-effort}.

\noindent \textbf{What tasks benefit the most from tools?} \quad
In general, tasks that cover multiple domains experience the highest increase, such as the ToolAlpaca benchmark in tool-using and the BigBench dataset in tool-making scenarios. Nonetheless, substantial gains may be expected
\begin{wrapfigure}[12]{r}{0.33\textwidth}
\vspace{-3mm}
\includegraphics[width=0.31\textwidth]{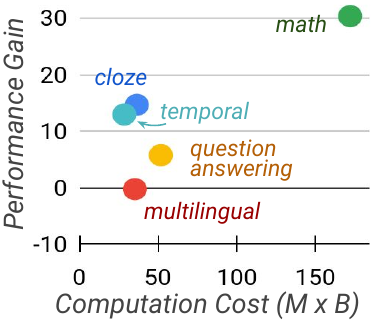}
\vspace{-3mm}
\caption{Compute \& performance gain with ToolFormer.}
\vspace{-2mm}
\label{fig:toolformer-cost}
\end{wrapfigure}
on API benchmarks (i.e., API-Bank and ToolAlpaca), because all examples are synthesized use cases for designated tools (\S\ref{sub:app-spec}), no-tool baselines are deprived of necessary components (i.e., tools) to solve the task, therefore achieving much lower accuracy.

On existing benchmarks, the ToolFormer method is the most efficient on MATH problems, showing the highest $30.4$ increase with little computation ($0.17$ MB). While other tasks improve less, multilingual tasks even degrade by $-0.2$ points, despite using a similar amount of compute. This variance across tasks aligns with expectations: using a \texttt{calculator} tool greatly improves the arithmetic ability of probabilistic LMs, which are not naturally suitable for symbolic calculations; however, LMs are originally built to solve language tasks such as machine translation (MT), so assigning the MT task to another (usually smaller) LM may not bring substantial improvements.

\begin{wrapfigure}[14]{r}{0.33\textwidth}
\vspace{-2mm}
\includegraphics[width=0.32\textwidth]{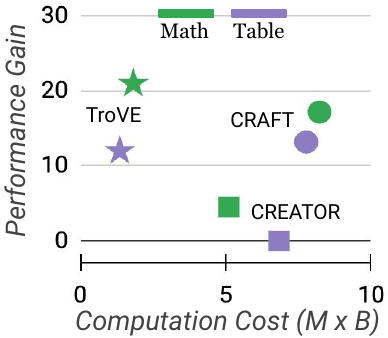}
\vspace{-3mm}
\caption{Comparing different tool-making methods.}
\vspace{-1mm}
\label{fig:tool-make-cost}
\end{wrapfigure}
\noindent \textbf{What methods are efficient in tool-making?} \quad
While it is hard to conduct fair comparisons for many works experimenting on different datasets, in tool-making scenarios (\autoref{fig:tool-make-cost}), the three methods (Creator, CRAFT, \textsc{TroVE}) experiment on the same MATH and TabMWP datasets, thus enabling fair comparisons in both cost and performance dimensions.
\textsc{TroVE} appears to be the most efficient method in general, costing only $1.2$--$1.4$K tokens while improving the performance by $12.0$--$21.0$ points in accuracy. 
In contrast, CREATOR and CRAFT are less efficient, costing $3.8$--$6.0$ times of compute, yet achieve only minimal ($0.0$--$4.5$\%) or comparable ($4.1$--$5.0$\%) accuracy increases.

\noindent \textbf{Training-time vs inference-time cost} \quad
Training-time and inference-time costs may not be equally important to many practitioners, since inference may be run many times but training often only needs to be done once.\footnote{Another measure of the inference process is latency, which also heavily depends on implementation or hardware choices. We do not report latency since these methods are implemented differently.} If we only consider inference-time cost in \autoref{tab:compute-cost}, the efficiency ranking of tooling methods changes. On one hand, tool-making method rankings roughly remain the same, except that CRAFT requires less compute than CREATOR on both tasks after getting rid of the training cost.
On the other hand, however, the ranking among tool-using methods drastically changes: ToolFormer requires more compute than API-Bank and ToolAlpaca when considering only inference-time cost. We conjecture this is mainly due to differences in baseline setups: ToolFormer adds in-context examples than the CoT baseline, API-Bank and ToolAlpaca use the same prompt for baseline and fine-tuned LMs with varied abilities to utilize tools presented in the prompt.
In general, if the user has sufficient budgets for training but higher demands on inference-time efficiency, the training approaches proposed by API-Bank and ToolAlpaca could be more suitable.

\section{Final Remarks}
Our survey provides definitions for LM-used tools and systematic summaries of existing approaches. While our empirical analysis guides when (on what tasks) and how (use what methods) should one use tools, we hope readers can more clearly understand the scenarios and techniques of LM tooling, from basic paradigm to advanced settings, and across LMs speaking natural and programming languages. 
% \dfcomment{I feel like the `what methods' part was comparatively underexplored -- and rightly so, since methods are still pretty new and haven't been thoroughly compared. Maybe we should instead focus, here, on the `how-to-evaluate' aspects which we were more concrete on?}
% Further, we point out missing aspects of tool evaluation and suggest more comprehensive measurements for efficiency, performance, reliability, and safety.
We believe tools can greatly extend and facilitate LM abilities, and hope our work elicits more discussions and research developments in %this direction.
(i) proposing benchmarks with natural use cases and executable tools, (ii) utilizing comprehensive evaluation metrics proposed in \S\ref{sec:good-tool}, and (iii) exploring more challenging and realistic scenarios for tool-using and tool-making techniques.
% (i) proposing better benchmarks, methods, and evaluation metrics, as well as (ii) advanced topics empowered by tools such as more capable agents and their interactions with humans or the world.
% \dfcomment{could consider broader directions for future work here enabled by your perspective, e.g. what other sorts of things might be implemented as tools that haven't been already? (interaction with people? reads and writes from an external memory?)}
% \gncomment{I agree with Daniel's comment above. I think that more insight into future directions or missing pieces would be nice here. Right now it's basically ``better evaluation'' and ``better methods'' which isn't super-insightful -- those are always the things that we can do :)}

% \subsubsection*{Author Contributions}
% If you'd like to, you may include a section for author contributions as is done
% in many journals. This is optional and at the discretion of the authors.

\subsubsection*{Acknowledgments}
We thank Saujas Vaduguru, Sherry Tongshuang Wu, Jiawei Liu, Shihao Liang, Pengfei Liu for the helpful discussions. Zora Zhiruo Wang is supported by the Teng Family Presidential Fellowship. Hao Zhu is supported by NSF EAGER Award \#2141751.

\bibliography{colm2024_conference}
\bibliographystyle{colm2024_conference}

\clearpage
\appendix
\section{Detailed computation process for tooling trade-offs}
\label{app:learning-effort}

\begin{figure}[ht]
\centering
\includegraphics[width=\textwidth]{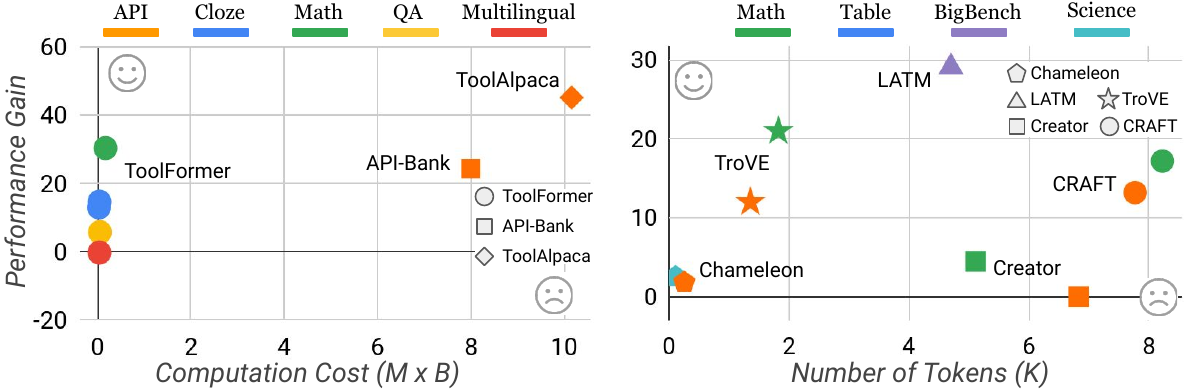}
\vspace{-6mm}
\caption{Computation cost of different approaches using open-source (left) and closed-source (right) models, and their performance gain on experimented datasets. We use different colors to represent tasks and different shapes to represent methods.}
% \vspace{-2mm}
\label{fig:overall-cost}
\end{figure}

For each method measured in \S\ref{sub:trade-off}, we describe the detailed processes in estimating their computation cost and performance improvement. 
For open-source models, we estimate cost $C = 6ND$, where $N$ is the number of tokens and $D$ is the parameter size (\autoref{fig:overall-cost}, left). Because the parameter size $D$ of closed-source GPT is unknown, we only measure the number of extra tokens $N$ per example (\autoref{fig:overall-cost}, right).

% \textbf{\begin{table}[ht]
% \small
% \begin{center}
%     \begin{tabular}{lllll}
%     \toprule
%     \multicolumn{1}{c}{\textbf{Work}} & \multicolumn{1}{c}{\textbf{Baseline}} & \multicolumn{1}{c}{\textbf{Tool-Use LM}} & \multicolumn{1}{c}{\textbf{Compute}} & \multicolumn{1}{c}{\textbf{Type}} \\
%     \midrule
%     {API-Bank} & {Alpaca} & {Lynx} & {3-epoch train} & {tool-use, train \& test} \\
%     {ToolAlpaca} & {Vicuna} & {ToolAlpaca} & {3-epoch train} & {tool-use, train \& test} \\
%     {Toolformer} & {GPT-J} & {Toolformer} & {25$k$ examples train} & {tool-use, train \& test} \\
%     {LATM} & {CoT} & {LATM} & {train, verify, test} & {tool-make, train \& test} \\
%     {CRAFT} & {PoT} & {CRAFT} & {train, verify, test} & {tool-make, train \& test} \\
%     % \midrule
%     {Chameleon} & {CoT/PoT} & {Chameleon} & {few-shot, verify} & {tool-use, test} \\
%     {CREATOR} & {PoT} & {CREATOR} & {abstract, decide, rectify} & {tool-make, test} \\
%     {TroVE} & {Primitive} & {TroVE} & {3-way generation} & {tool-make, test} \\
%     \bottomrule
%     \end{tabular}
% \end{center}
% \caption{Details for measuring computation cost and task performance.}
% \label{tab:est-compute-cost}
% \end{table}}

\subsection{Methods using known-sized models}

For methods using models whose parameter sizes are known, we estimate the computation cost by the FLOPs during any additional modules such as training and inference with additional context. In general, the computation cost is majorly affected by (1) the number of tokens processed, and (2) the parameter size of models.

\noindent \textbf{API-Bank \citep{li-etal-2023-api}} \quad
This work trains the Lynx model that uses tools to solve problems in the proposed API-Bank dataset. The Lynx model is initialized by Alpaca 7B parameters, and trained on the API-Bank training set with 3 epochs. Therefore, we adopt the Alpaca 7B as the baseline and Lynx as the tool-using model, where the 3-epoch training is the additional computation cost introduced to enable tool use.
We calculate the total number of tokens involved in the training process, including the example i/o and additional instructions. 
Because the baseline and proposed method use the same prompt at inference time, no additional computation is required.
Regarding task performance, we adopt the total correctness across all evaluation systems, as reported in Table 3. We report the difference between the fine-tuned Lynx-7B and the zero-shot Alpaca-7B.

\noindent \textbf{ToolAlpaca \citep{tang2023toolalpaca}} \quad
This work proposes the ToolAlpaca dataset and trains Vicuna models to use tools. The baseline models are Vicuna-7B and Vicuna-13B models. The trained tool-using models are called ToolAlpaca-7B and ToolAlpaca-13B models. All ToolAlpaca models are trained on the training split for 3 epochs, so we estimate the cost during this training process for 7B and 13B models, respectively.
We adopt the `overall' results reported in Table 3, on examples with both simulated tools and real-world APIs, and report their average results. We measure the performance gain by the difference between the ToolAlpaca-7/13B and Vicuna-7/13B.

\noindent \textbf{Toolformer \citep{schick2023toolformer}} \quad
This work integrates five tools --- question answering system, calculator, Wikipedia search, machine translation system, and calendar --- respectively for five tasks transformed from a subset of CCNet \citep{wenzek2020ccnet}.
Starting with GPT-J models \citep{wang2021gpt-j} as the no-tool baseline, they train on 25$k$ model-synthesized examples for each tool and obtain the Toolformer models, causing a total of 1$M$ FLOPs for each task. At inference time, they add special instruction and in-context examples to prompt tool using, resulting in extra compute.
Because each task contains multiple datasets, we report the average results to represent the general task performance.

% ######################
\subsection{Models with unknown size}

While many of the works use GPT-3.5 or GPT-4 models that do not release their parameter size, we estimate the cost by using the number of tokens processed in extra modules.

\noindent \textbf{Chameleon \citep{lu2023chameleon}} \quad 
This work proposes to take a tool-augmented approach to improve on two existing datasets --- ScienceQA and TabMWP. Because all experiments use ChatGPT and GPT-4 models, whose parameter sizes are unknown, we only examine results with (the better) GPT-4 model to fairly compare with other methods using GPT-4. 
Specifically for the ScienceQA dataset, we adopt the Chain-of-Thought (CoT) baseline reported in the paper, and report task accuracy as in the \textsc{All} column in Table 3. We calculate the difference in number of tokens between the proposed Chameleon methods against the CoT baseline.
For the TabMWP dataset, we adopt the Program-of-Thought (PoT) baseline and similarly calculate the token number difference using the provided results.\footnote{\url{https://github.com/lupantech/chameleon-llm}} We adopt numbers in the \textsc{All} column in Table 4 as the TabMWP accuracy.

\noindent \textbf{LATM \citep{cai2023large}} \quad
This work proposes to use LMs to make tools for individual tasks in BigBench. Compared to the chain-of-thought (CoT) baseline, the proposed LATM method integrates training, validation, and inference stages to make tools and solve questions. We estimate the compute cost by the additional number of tokens used for LATM than for CoT.
We measure each method by averaging its accuracy across all six selected tasks.

\noindent \textbf{CRAFT \citep{yuan2023craft}} \quad
This work uses LMs to make tools for math, table, and image reasoning tasks. We calculate the number of tokens used during training and inference, using its released code and data.\footnote{\url{https://github.com/lifan-yuan/CRAFT}}
CRAFT similarly implements CoT as the baseline, and proposes further training, verification, and finally testing in the CRAFT method. 
We report its task accuracy on the representative datasets from each task --- MATH, TabMWP, and GQA --- to enable fairer comparison with other works having overlapping datasets.

\noindent \textbf{CREATOR \citep{qian2023creator}} \quad
As a prior work for CRAFT, CREATOR similarly tests on MATH and table tasks, but designs its methods differently. In addition to CoT, this work implements a stronger program-oriented baseline called Program-of-Thought (PoT). We also adopt PoT as the main baseline without tool making or using. The CREATOR method operates at test time, with multiple steps through tool making, solution generation, verification, rectification, etc. We calculate the difference in number of tokens between the CREATOR approach and the baseline PoT setting.
We adopt the task accuracy reported in Table 2 (MATH) and Table 3 (TabMWP) from the original paper.

\noindent \textbf{TroVE \citep{wang2024trove}} \quad
TroVE also induces tools without training supervision. This work adopts the primitive baseline, a presumably stronger version of PoT yet without much textual explanation. The main implementation change in TroVE is the three-mode generation and multi-candidate sampling. We calculate the additional tokens used in TroVE compared to the primitive baseline.
The dataset reports task accuracy, solution complexity, and toolbox size, we only adopt the task accuracy to fairly compare with other works.

% \begin{figure}[ht]
% \centering
% % \vspace{-1mm}
% \includegraphics[width=\textwidth]{./figures/inference-cost.pdf}
% \vspace{-6mm}
% \caption{Performance gain versus inference-time computation cost.}
% \vspace{-3mm}
% \label{fig:inference-cost}
% \end{figure}

% \begin{table}[ht]
% \small
% \begin{center}
% \resizebox{0.92\linewidth}{!}{
%     \begin{tabular}{ll}
%     \toprule
%     \multicolumn{1}{c}{\textbf{Missing aspects}} & \multicolumn{1}{c}{\textbf{Potential metrics}} \\
%     \midrule
%     {Efficiency of tool integration} & {computation overhead} \\
%     {Quality of tools} & {runtime and memory usage of tools} \\
%     {Reliability of unstable tools} & {success rate of tools} \\
%     {Reproducible testing} & {execution correctness w.r.t. canonical trajectory} \\
%     {Safe usage} & {visibility, data integrity, and other meta-data about tools} \\
%     \bottomrule
%     \end{tabular}
%     }
% \end{center}
% \caption{Missing evaluation aspects and potential metrics.}
% \label{tab:missing-eval}
% \end{table}
% \input{appendix/embodied}

\end{document}